\documentclass[]{article}
\usepackage{aclsub}
\usepackage{epsfig}

\title{Generating a 3D Simulation of a Car Accident from a Written Description in Natural Language: the CarSim System}
\author{Sylvain DUPUY, Arjan EGGES, Vincent LEGENDRE, \and Pierre NUGUES\\ \\
        GREYC laboratory\\
        ISMRA\\
        6, bd du Mar\'{e}chal Juin\\
        F-14050 Caen, France\\
        Phone: (33) 231 452 705\\
        Fax: (33) 231 452 760\\
        Email: \{dupuy,vlegendr\}@ensicaen.ismra.fr\\pnugues@greyc.ismra.fr\\egges@cs.utwente.nl}
\summary{This paper describes a prototype system to visualize and animate
3D scenes from car accident reports, written in French. The
problem of generating such a 3D simulation can be divided into two
subtasks: the linguistic analysis and the virtual scene
generation. As a means of communication between these two modules,
we first designed a template formalism to represent a written
accident report. The \textsc{CarSim} system first processes
written reports, gathers relevant information, and converts it
into a formal description. Then, it creates the corresponding 3D
scene and animates the vehicles.
} 
\keywords{information extraction, visual simulation, space and
time} \contact{dupuy@ensicaen.ismra.fr} \conference{no}

\begin{document}
    \newcommand{\obj}[1]{\textit{#1}}
    \newcommand{\Carsim}{\textsc{CarSim}}
    \newtheorem{definition}{Definition}
    \hyphenation{collisions}

    \authorname{Sylvain DUPUY, Arjan EGGES, Vincent LEGENDRE, \and Pierre NUGUES}


    \maketitle

    \begin{abstract}
        This paper describes a prototype system to visualize and animate
3D scenes from car accident reports, written in French. The
problem of generating such a 3D simulation can be divided into two
subtasks: the linguistic analysis and the virtual scene
generation. As a means of communication between these two modules,
we first designed a template formalism to represent a written
accident report. The \textsc{CarSim} system first processes
written reports, gathers relevant information, and converts it
into a formal description. Then, it creates the corresponding 3D
scene and animates the vehicles.

    \end{abstract}
    \section{Introduction}

This paper describes a prototype system to visualize and animate a
3D scene from a written description. It considers the narrow class
of texts describing car accident reports. Such a system could be
applied within insurance companies to generate an animated scene
from reports written by drivers. The research is related to the
TACIT project \cite{bib:enj96} at the GREYC laboratory of the
University of Caen and ISMRA.

There are few projects that consider automatic scene generation
from a written text, although many projects exist that incorporate
natural language interaction in virtual worlds, like Ulysse
\cite{bib:ber98,bib:god99} or AnimNL \cite{bib:anim93}.
Visualizing a written car accident report requires a different
approach. It is closer to projects focusing on text-to-scene
conversion, like WordsEye \cite{bib:wordseye}. However, unlike the
latter, our objective is to build an animation rather than a
static picture and behavior of dynamic objects must then be taken
into account. There also exist systems that carry out the reverse
processing, from video data to text description, as ANTLIMA
\cite{bib:schirraantlima},~\cite{bib:blocher95optional}.

\begin{figure}[htbp]
 \begin{center}
\epsfig{file=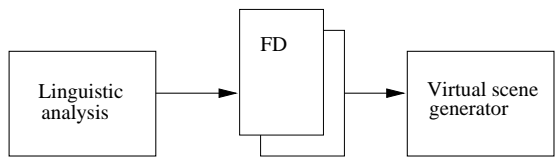,height=2cm}  
 \end{center}
 \caption{The two
subsystems and the FD (Formal Description) as a means of
communication.} \label{fig:twosubsystems}
 \end{figure}

We present here an overview of the \textsc{CarSim} system that
includes a formalism to describe and represent car accidents, a
linguistic module that summarizes car accident reports according
to this formalism, and a visualizing module that converts formal
descriptions to 3D animations. In our case, the linguistic module
has to deal with texts where syntax and semantics involve time and
space description and simultaneous actions of two or more actors
(i.e. the cars). Figure~\ref{fig:twosubsystems} illustrates the
whole architecture of the system. The visualizing module is
extensively described in \cite{bib:stageverslag}.

The remainder of our paper is organized as follows.
Section~\ref{Formal representation} presents the formalism for
describing an accident. Section~\ref{Extraction} describes the
template filling methods that lead to the conversion of a text
into its formal representation. Section~\ref{Planning} covers
planning techniques and accident modelling algorithms that we use.
Finally, Section~\ref{Simulating} presents and discusses the
evaluation of the system on the test corpus (MAIF corpus).

\section{Formal Representation in \textsc{CarSim}}
\label{Formal representation}

\small
\begin{quotation}
``V\'{e}hicule B venant de ma gauche, je me trouve dans le
carrefour, \`{a} faible vitesse environ 40 km/h, quand le
v\'{e}hicule B, percute mon v\'{e}\-hi\-cu\-le, et me refuse la
priorit\'{e} \`{a} droite. Le premier choc atteint mon aile
arri\`{e}re gauche, sous le choc, et \`{a} cause de la
chauss\'{e}e glissante, mon v\'{e}hicule d\'{e}rape, et percute la
protection m\'{e}tallique d'un arbre, d'o\`{u} un second choc
frontal.'' \emph{Text A4, MAIF corpus.}

``I was driving on a crossroads with a slow speed, approximately
40 km/h. Vehicle B arrived from my left, ignored the priority from
the right and collided with my vehicle. On the first impact, my
rear fender on the left side was hit and because of the slippery
road, I lost control of my vehicle and hit the metallic protection
of a tree, hence a second frontal collision.'' \emph{Text A4, MAIF
corpus, our translation.}
\end{quotation}
\normalsize

The text above is an accident report from the
MAIF\footnote{Mutuelle Assurance Automobile des Instituteurs de
France. MAIF is a French insurance company.} corpus, which
contains 87 texts in French. It is a good example of the possible
contents of an accident description: a rather complex interaction
between a set of different objects (two cars and a tree). This
section describes the formal representation used in the
\textsc{CarSim} system. The example of Text A4 will be explained
with more details in Section~\ref{example}.

\subsection{The General Accident Model}

In \textsc{CarSim}, the general accident model consists of three
lists of objects: motionless objects (\mbox{STATIC}), moving
objects (\mbox{DYNAMIC}), and finally collisions
(\mbox{ACCIDENT}).

\mbox{STATIC} and \mbox{DYNAMIC} lists describe the general
environment in which the accident takes place. Knowing them, the
accident itself is the only remaining item to determine. Using
manual simulation, we realized that most accidents in the corpus
could be framed using an \emph{ordered list of
collisions}\footnote{Two collisions will never happen at the same
time.}. Each collision is represented by a relation between two
objects either in \mbox{DYNAMIC} and/or \mbox{STATIC}
lists

\subsection{Static Objects}

In general, a static object can be defined with two parameters:
one defining the nature of the object and another one that defines
its location. In \textsc{CarSim}, a static object can be either a
road type or an object that can participate in a collision (e.g. a
tree). In the formal description, a reference to the latter kind
of object can occur in a collision specification. This is why
these static objects are defined with an identity parameter (ID).

Concerning \mbox{ROAD} objects, their nature is specified in the
\mbox{KIND} parameter. The possible KIND values in the present
prototype are: \emph{crossroads}, \emph{straightroad},
\emph{turn\_left}, and \emph{turn\_right}.

\mbox{TREE}s, \mbox{LIGHT}s (traffic lights), \mbox{STOPSIGN}s,
and \mbox{CROSSING}s (pedestrian crossings) are the other possible
static objects. Their location is given by the \mbox{COORD}
parameter. Since trees and traffic lights can participate in
collisions, they also have an \mbox{ID}, that allows further
references. Finally, traffic lights contain a \mbox{COLOR}
parameter to indicate the color of the light (\emph{red},
\emph{orange}, \emph{green} or \emph{inactive}).

\subsection{Dynamic Objects}



Dynamic objects cannot be defined by giving only their nature and
position. Rather than the position, the \emph{movement} of the
object must be defined.

In the \textsc{CarSim} formal representation, each dynamic object
is represented by a \mbox{VEHICLE}, with a \mbox{KIND} parameter
indicating its nature, (\emph{car} or \emph{truck}) and a unique
identifier \mbox{ID}. The movement of a dynamic object is defined
by two parameters. The \mbox{INITIAL DIRECTION} defines the
direction to which the object is headed before it starts driving
(\emph{north}, \emph{south}, \emph{east}, or \emph{west}). The
second parameter is an ordered list of atomic movements that are
described by \mbox{EVENT}s. This list is called the \emph{event
chain} and corresponds to the \mbox{CHAIN} parameter. \mbox{KIND}
specifies the nature of each event. At present, \textsc{CarSim}
recognizes the following events: \emph{driving\_forward},
\emph{stop}, \emph{turn\_left}, \emph{turn\_right},
\emph{change\_lane\_left}, \emph{change\_lane\_right},
\emph{overtake}.

\begin{figure}[htbp]
 \begin{center}
 \epsfig{file=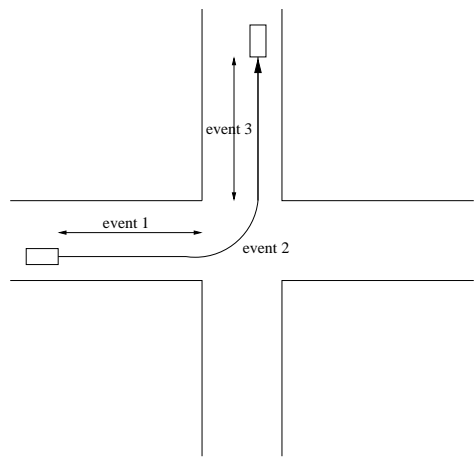,height=6cm}\\
\end{center}
 \caption{A crossroads with a vehicle driving forward, turning
left and driving forward with an
 initial direction to the East.}
\label{fig:eventchain}
 \end{figure}

With these definitions, we could define a dynamic object with
$\mbox{KIND} = \obj{car}$, $\mbox{INITIAL DIRECTION} = \obj{East}$
and $\mbox{CHAIN} = < $\obj{driving\_forward, turn\_left,
driving\_forward}$>$. Figure~\ref{fig:eventchain} shows the motion
of this vehicle.

\subsection{Collisions}

As we said before, the accident is described by an ordered list of
collisions. The order of the collisions in the list corresponds to
the order in which they take place in the accident simulation. A
collision is defined by giving the two objects that participate in
the collision and some additional attributes. At present, these
attributes are the collision coordinates and the parts of the
vehicles that are involved in the collision (participating parts).

There is a slight distinction between the vehicle that collides
(in other words: the \emph{actor}) and the vehicle that is hit
(the \emph{victim}). For planning reasons (and also for linguistic
grounds) it is useful to maintain this distinction in the
formalism. To summarize, a collision occurs between an
\emph{actor} and a \emph{victim}. The victim can be either a
static or a dynamic object, the actor clearly has to be a dynamic
object. The notions of \emph{actor} and \emph{victim} are not
related with the responsibility of one particular vehicle within
the accident. This kind of relationships must be deduced from a
complex responsibilities analysis, that could be based on the
traffic rules.

Next to the location (coordinates) of the collision, something has to be said
about the configuration of the
 objects while colliding. The participating parts are
sometimes given in the text, see for example Text A4 at the
beginning of this section. The \textsc{CarSim} system uses a
simplified model of these vehicle parts. They are divided in four
categories: \emph{front}, \emph{rear}, \emph{leftside}, and
\emph{rightside}, plus one \emph{unknown} category.

\subsection{An Example}
\label{example}

In order to give an example of a formal accident description and
also to introduce the linguistic part, we will give now more
details about the manually written FD of Text A4.

In a written text, information can be given either explicitly or
implicitly. Besides, the contents of implicit information differs
in each text. In Text A4, what information can we directly gather
from the text?

Text A4 describes an accident with two collisions, involving two
vehicles and a tree. It takes place at a crossroads. The first
collision involves two vehicles. One of them is referred to as the
``vehicle B'', the other is the narrator's vehicle (``my
vehicle''). From now on, vehicles will be called
\emph{\mbox{vehicleB}} and \emph{\mbox{vehicleA}} respectively.
The second collision involves \emph{\mbox{vehicleA}} and the tree.
In the FD, the tree is identified in a unique way as \emph{tree1}.
From this information, we already know how many objects will be
needed to describe the accident: two static objects (a crossroads
and a tree \emph{tree1}), two dynamic objects
(\emph{\mbox{vehicleB}} and \emph{\mbox{vehicleA}}) and finally
two collisions.

The text does not mention any special behavior of the two
vehicles. They are both driving when the accident occurs. Hence,
the event chain is the same for both vehicles, a single
\emph{driving\_forward} event.

The roles played by the vehicles in each collision are also given.
As human beings, we deduce them from the grammatical functions of
the noun groups or pronouns referring to the vehicles in the
sentences where collisions are described. In the first collision,
the actor is \emph{\mbox{vehicleB}} and the victim
\emph{\mbox{vehicleA}} (respectively, subject and object of the
verb ``percuter'', ``to collide with'' in the translation). In the
second one, the actor is \emph{\mbox{vehicleA}} and the victim
\emph{\mbox{tree1}}.

The parts of the vehicles that participate in a collision are
sometimes explicitly given in the report, as for example for
\emph{\mbox{vehicleA}} in Text A4. In the first collision, the
impact occurs at the rear left-hand side of the vehicle (``On the
first impact, my rear fender on the left side was hit'') and in
the second one, \emph{\mbox{vehicleA}} hits the tree with the
front of the car (``hence a second frontal collision'').\\

Actually, we don't know whether the vehicles in the text are cars,
trucks or something else. As no precise information is explicitly
given in the text, we simply assume that these vehicles are
cars\footnote{\emph{car} will be the default value of the KIND
parameter of dynamic objects.}. The type of vehicles is not the
only implicit piece of information in the text. The initial
directions of the vehicles are only known relatively to each
other. We know that \emph{\mbox{vehicleB}} is coming from the
left-hand side of \emph{\mbox{vehicleA}} (``Vehicle B arrived from
my left'') and if we arbitrary decide that \emph{\mbox{vehicleA}}
starts heading to the North, then \emph{\mbox{vehicleB}} has to
start heading to the East. The same fragment of the text gives us
the participating part of \emph{\mbox{vehicleB}}. Since the
participating part of \emph{\mbox{vehicleA}} in the first
collision is \emph{leftside}, we can conclude that
\emph{\mbox{vehicleB}}'s part is \emph{front}. The tree has no
particular participating part. Thus, it will be defined as
\emph{unknown} but we can assume that the impact occurs with the
trunk because all the scene takes place in a two-dimensional
plane.

Below is the formal description of Text A4 that can be given to
the simulation module of \textsc{CarSim}:

\pagebreak[3]

\footnotesize
\begin{verbatim}
// Static objects
STATIC [
   ROAD [
      KIND = crossroads;
   ]
   TREE [
      ID = tree1; COORD = ( 5.0, -5.0 );
   ]
]

// Dynamic objects
DYNAMIC [
   VEHICLE [
      ID = vehicleB; KIND = car;
      INITDIRECTION = east;
      CHAIN [
         EVENT [
            KIND = driving_forward;
         ]
      ]
   ]
   VEHICLE [
      ID = vehicleA; KIND = car;
      INITDIRECTION = north;
      CHAIN [
         EVENT [
            KIND = driving_forward;
         ]
      ]
   ]
]

// Collision objects
ACCIDENT [
   COLLISION [
      ACTOR = vehicleB, front;
      VICTIM = vehicleA, leftside;
      COORD = ( 1.0, 1.0);
   ]
   COLLISION [
      ACTOR = vehicleA, front;
      VICTIM = tree1, unknown;
   ]
]
\end{verbatim}
\normalsize

The only information we did not discuss yet are the coordinates of
static objects and impacts. Coordinates are numbers. They are
never explicitly given in the text and obviously, even if some
numbers appeared in the text, the semantic of these numbers would
be implicit too. \textsc{CarSim} assumes that coordinates (0,0)
are the center of the scene. In Text A4, the origin is the center
of the crossroads. The first collision occurs in the crossroads,
hence the coordinates will be close to the origin. The coordinates
of the tree are chosen so that they match the idea of the scene as
a reader could imagine it. They also depend on the size of the
graphical objects that are used in the 3D scene (e.g. the size of
the roads).


    \section{The Information Extraction Task}
\label{Extraction}

The first stage of the \textsc{CarSim} processing chain is an
information extraction (IE) task that consists in filling a
template corresponding to the formal accident description (FD)
described in Section~\ref{Formal representation}. Such systems
have been already implemented, as FASTUS \cite{bib:fastus}, and
proved their robustness. Our information retrieval subsystem is
restricted to car accident reports and is goal-driven. The main
idea is to start from a default description, a pre-formatted
\mbox{FD}, that the IE task alters or refines using inference
rules. Hence, the default output will be a well-formatted
\mbox{FD}, describing a collision between two cars, even if the
given text is a poem.

\subsection{Parsing}

The first step of the information extraction process is a lexical
analysis and a partial parsing. The parser generates tokenized
sentences, where noun groups, verb groups, and prepositional
groups are extracted. The parser uses DCG rules \cite{bib:dcg} and
a dictionary containing all the words that occur in the corpus.


\subsection{Extracting Static Objects}

The formalism describes two types of static objects: the type of
road (the road configuration) and some other static objects (stop
signs, traffic lights, pedestrian crossings and trees). The method
used to extract these objects consists in looking up for keywords
in the tokenized text.

The extraction of static objects is done at the beginning of the
information extraction task. We realized that the road
configuration is the most relevant piece of information in the
description of an accident, since it conditions all the following
steps (see Section~\ref{Generating Collisions} for further
explanations).

The formalism considers four different configurations:
\emph{straightroad}, \emph{crossroads}, \emph{turn\_left}, and
\emph{turn\_right}. In the present system, we restricted it to
three types of road:
\begin{itemize}
\item \emph{crossroads}, indicated by cue words such as ``carrefour'',
``intersection'', ``croisement'' (crossroads, intersection,
junction).
\item \emph{turn\_left}, with cues such as ``virage'', ``courbe'', ``tournant'' (bend, curb, turn). We assume that
\emph{turn\_left} and \emph{turn\_right} are equivalent.
\item \emph{straightroad}, that corresponds to the situation when none of the previous words have been
found.
\end{itemize}

\subsection{Extracting Collisions}

A collision consists of a verb, an actor, a victim and of the
participating parts of the two vehicles. We select verbs
describing a collision such as ``heurter'' (``to hit''), ``taper''
(``to bang''), ``percuter'' (``to crash into''), ``toucher'' (``to
touch''),\ldots

For each extracted verb, the system checks whether the verb group
is in passive or active form, then identify the related
grammatical relations: subject-verb and verb-object or verb-agent.
Extraction techniques of such dependencies have already been
implemented, as in \cite{bib:chanod}. Our system uses three
predicates in order to find the subject (\emph{find\_subject}) and
either the object (\emph{find\_object}) or the agent
(\emph{find\_agent}) of the verb. If the verb is in an active
form, it makes the assumption that the subject and the object of
the verb will be respectively the actor and the victim of the
collision. In the case of a passive form, the subject will be the
victim and the agent, the actor.

Below is the sketch of the algorithm of these three predicates:
\begin{itemize}
\item \emph{find\_subject} looks for the last noun group before
the verb that describes a valid actor, that is a vehicle or a
personal pronoun like ``je'' (``I''), ``il'' (``he''), or ``nous''
(``we'') .
\item \emph{find\_object} starts looking for the first noun group
after the verb that describes a valid victim, that is both
vehicles and static objects. If no valid victim is found, it
searches for a reflexive or personal pronoun inside the verb
group. In case of failure, the first noun group after the verb is
chosen.
\item \emph{find\_agent} looks for a valid actor in a
prepositional group introduced by ``par'' (``by'').
\end{itemize}

\subsection{Generating Collisions and Dynamic Objects}
\label{Generating Collisions}

For each collision, the system tries to extract the participating
parts of the vehicles in the noun groups that refer to the actor
and the victim. To do this, it looks for cues like ``avant'',
``arri\`ere'', ``droite'', or ``gauche'' (``front'', ``rear'',
``right'', or ``left'').

Then, the system creates two dynamic objects (see
Section~\ref{dumping}) and a collision between them. The generated
properties of the collision depend on the road configuration:
\begin{itemize}
\item Straight road: the first vehicle heads to the East, the
other one starts from the opposite end of the road, heading to the
West. The collision is a head-on impact.
\item Turn: The first vehicle starts heading to the East, then turns to the Left.
The second one starts heading to the South, then turns to the
Right. The collision is frontal and happens at the center of the
turn.
\item Crossroads: We choose to represent here the most
frequent traffic offence (in France). The first vehicle drives
straight to the East, the second one drives to the North. The
front of the actor's vehicle collides with the left-hand side of
the victim.
\end{itemize}

As we do not extract the initial directions of the vehicles, these
three cases are the only possible ones. When the system cannot
find the actor or the victim of a collision, default objects are
created matching the road configuration.

\subsection{Deleting Useless Objects}
\label{dumping}

When creating collision objects, two new vehicles are instantiated
for each collision, even if the victim is a static object.
Moreover, one vehicle can obviously participate in several
collisions. All the unnecessary vehicles should then be thrown
away.

A vehicle that represents a static object can be removed easily,
since the real static object still exists. All we have to do is to
modify the reference given in the victim parameter of the
collision in the template, then delete the redundant vehicle.

Deleting the duplicates is more difficult and involves a
coreference resolution. An identification mechanism of the
narrator has been added to the system. All the personal pronouns
in the first person or some expressions like \mbox{``the vehicle
A''} will be designated with the id \emph{enunciator}. In the
other cases, coreference occurs only when the two ids are strictly
the same (in the sense of string comparison). Then, the system
keeps only the first created object between the duplicates and
delete the others.

\subsection{Extracting Event Chains}

The vehicles generally do not drive straight forward. They carry
out two or more successive actions. In the formal description,
these possible actions correspond to the events of dynamic objects
and are in limited number: \emph{\mbox{driving\_forward}},
\emph{\mbox{turn\_left}}, \emph{\mbox{turn\_right}},
\emph{\mbox{change\_lane\_right}},
\emph{\mbox{change\_lane\_left}}, \emph{overtake}, and
\emph{stop}.

In written reports, these actions are mostly indicated by verbs.
The system has to identify them and to link the corresponding
event(s) to the appropriate vehicle. When the subject is
identified as the narrator, the link is obvious. In the other
cases, if there are only two vehicles, the narrator and another
one, a new event is added to the event chain of the second
vehicle. Otherwise, the system checks whether the subject of the
verb is strictly identical (string comparison) to one vehicle's
id. In this case, a new event is also created and added to the
event chain. Some verbs imply multiple events, e.g.
``red\'emarrer'' (``to get driving again'') that indicates that
the driver stopped beforehand. Consequently, a \emph{stop} event
then a \emph{driving\_forward} event are added.

With this simple extraction mechanism, the order of the events in
the event chain does not necessarily respect the chronology but
rather the order of the text. We assume that the story is linear,
which is the case in most accident reports.

\subsection{Writing the Formal Description}

The final step of the linguistic part consists in formatting a
template corresponding to the accident description. Because the
inferred facts have exactly the same attributes as the formalism's
elements, a very simple transcription algorithm is used to convert
the facts in a text file that can be processed afterwards by the
simulator.

    \section{Planning}
\label{Planning}

Planning complex events like collisions requires a well-defined
and flexible planning architecture. General planning algorithms
which apply methods incorporating artificial intelligence, are
discussed in \cite{bib:nil98}. 
The \textsc{CarSim} planner is much more straightforward, because
the planning process is not as complex as a lot of traditional AI
planning problems, see also \cite{bib:nor95}. The total planning
process is performed by using five different subplanners, which
all perform a small part of the total planning task.

\subsection{The Preplanner}
The preplanner is a planner that ensures the consistency of the
formal description. If some values are not given (e.g. coordinates
of a static object or initial directions of dynamic objects) or
some values imply a contradiction (a vehicle turning left on a
straight road), this planner tries to find (default) values and to
solve the conflicts. This planner is a simple knowledge base, as
discussed in \cite{bib:nor95}.

\subsection{The Position Planner}
The position planner estimates the start and end positions of the
vehicles in the simulation. By default, a vehicle is placed 20
meters away from the center of the (cross)road. If two or more
vehicles are moving in the same direction, they can't all be
placed at this distance because they would overlap. Therefore, if
there is more than one vehicle facing a particular direction, the
second vehicle is placed at a distance of 26 meters from the
center and if there is a third vehicle, it is placed at 32 meters
from the center\footnote{In the \textsc{CarSim} system, the
maximum number of vehicles that can have the same initial
direction is \emph{three}.}. Regarding the end points of the
vehicles, the vehicle that is placed closest to the center, will
have its end point placed farther away from the center. The
vehicle initially having a start point far away from the center
will have an end point close to the center, so that every vehicle
traverses approximately the same distance.

\subsection{The Trajectory Planner}
Based on the (very global) description of the movement of every
vehicle in the formal model, this planner constructs a trajectory,
represented by a set of points in the Euclidian space.
 Every event in the event chain is converted
to a list of trajectory points. A turn is approximated by a number
of points lying on a circle arc. Overtaking is modelled by using a
goniometrical function.

\subsection{The Accident Planner}
The accident planner uses the trajectory that is created by the
trajectory planner. Since event chains only include atomic
movements and not collisions, this trajectory is planned as if
there was no collision at all. The task of the accident planner is
to change this trajectory in such a way that it incorporates the
collision. Some part of it has to be thrown away and an
alternative part (which ultimately leads to the point of
collision) has to be added to the trajectory. For every vehicle,
actor or victim, the trajectory is thus changed in two steps:

\begin{enumerate}
\item
Remove a part of the trajectory.
\item
Add a part to the trajectory so that the final result will be a
trajectory that leads the vehicle to the point of collision.
\end{enumerate}

The part of the trajectory that has to be removed depends on the
coordinates where the collision occurs. We designed an algorithm
that draws a circle around the collision point and removes the
trajectory part that lies within the circle region. Also, the
segment that comes after the removed trajectory part is thrown
away, because a trajectory does not allow gaps. The radius of the
circle is thus a parameter that defines the precision of the
algorithm. If a large radius is chosen, a large part of the
trajectory will be removed. An application of the algorithm using
a small radius only removes the trajectory part closest to the
collision point.

\subsection{The Temporal Planner}
The temporal planner of the \textsc{CarSim} system is not a
planner in the sense of the planners described in \cite{bib:nil98}
The temporal planner of the \textsc{CarSim}
system plans the temporal values of the trajectory in two steps.
Generally, a trajectory consists of a number of `normal'
trajectory points, followed by a number of trajectory
 points that represent a collision. First the segment that is not part of any collision
is planned. After that, the system plans the remaining segment. In
the \textsc{CarSim} system, every trajectory point has a
\emph{time value}. This is a value between 0 and 1, with 0
representing the beginning of the simulation and 1 being the end
of it. The temporal planner tries to find time values for the
trajectory points so that the collisions happen in a natural way.

    \section{Results and Discussion}
\label{Simulating}

The \textsc{CarSim} system has been implemented and evaluated over
the MAIF corpus. The assessment method does not consist, as
usually done with IE systems, in calculating a precision and a
recall. Our objective is to design a system that carries out the
whole processing chain, that is from a written report up to a 3D
animation. Therefore, we preferred to compare the simulation with
the understanding and mental representation of the scene that
could have a human reader. This implies that some aspects of the
formal description are not taken into account when evaluating the
system, e.g. we assume that the value of the INITIAL\_DIRECTION
parameter is less important than the positions of the vehicles
relatively to each other. Hence, we considered that the result is
acceptable as far as the latter is correct.

\begin{figure}[h]
\begin{center}
\epsfig{file=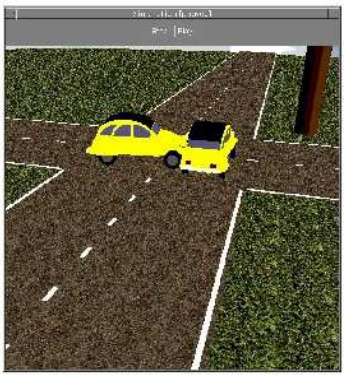,height=5cm}\\
\end{center}
\caption{The first collision in Text A4.} \label{fig:firstcolA4}
\end{figure}

\begin{figure}[h]
\begin{center}
\epsfig{file=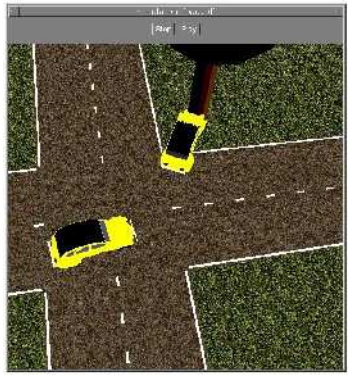,height=5cm}\\
\end{center}
\caption{The second collision in Text A4.} \label{fig:seccolA4}
\end{figure}

According to such criteria, we considered that the simulation
provided by the system corresponds, in 17\% of the texts, with
what could have imagined a human being.
Figure~\ref{fig:firstcolA4} \&~\ref{fig:seccolA4} show the two
collisions described in Text A4.

Failure cases have many different grounds. They may be related
either to the IE task, to the simulator, or to a lack of
cooperation between the two subsystems. 
Evaluating separately each subsystem leads to a better
understanding of the actual limits of the system.

Feeding the simulator with manually written formal descriptions
provides a good way to evaluate it for itself. According to such
tests, the \textsc{CarSim} system generates an acceptable
simulation of almost 60\% of the reports. This implies that the
results of the overall system will be lower. \textsc{CarSim}'s
simulator does not succeed in simulating manually written formal
descriptions because of three main causes: expressivity of the
formalism that does not cover all possible accidents (e.g.
synchronization between event chains of different objects), the
restricted number of scenarios considered by the \textsc{CarSim}
visualizer and the limited database of 3D graphical objects.
Depending on the text, the failure is the result of either only
one of these restrictions or a combination. Future work on the
project will focus on these issues.


The efficiency of the IE task varies with the nature of extracted
information. First, the results clearly depend on the accuracy
with which the system can correctly extract impacts, that is find
the verb representing the collision and also resolve the actor,
the victim and possibly their participating parts\footnote{when
the parts are explicitly described}. This task is successfully
accomplished in 69\% of the texts\footnote{In the rest, it
generates default impacts or impacts are erroneous.}. In addition,
the system correctly extracts EVENTS in 35\% of the texts. This
means that in 35\% of the texts, all the events are properly
extracted with a good ordering.\\


Concerning time and space information, the system provides only
simple mechanisms to obtain them. Our system is at an early stage
and our objective when designing it was to see whether such an
approach was feasible. It represents a sort of improved baseline
with which we can compare further results.
At this time, the temporal information known by the system is
restricted to the events associated with dynamic objects. Our
method assumes that they are given in the text in the same order
they occur in reality. This is a simplification that proves wrong
in some reports. Further improvements could take into account
tenses of verbs, temporal adverbs and prepositions, so that the
system could determine the real chronological relationships
between events.

A similar comment can be given with regards to spatial
information. In \textsc{CarSim}, the spatial configuration (the
background of the scene) is given mainly by the type of roads. The
extraction of participating parts also provides additional
information that influence the relative positions of the vehicles
when colliding. During preplanning stage, the system checks the
consistency of the FD and tries to resolve conflicts between the
different information. At present, initial directions of the
vehicles depend only on the background of the scene, that is the
road configuration. The coordinates are also chosen arbitrary from
the beginning. See for example the tree referred as \emph{tree1}
in Text A4: no information about its location is given in the
text. The only facts relative to it that we can deduce from the
original report are its existence and its involvement in a
collision. Moreover, the problem of choosing a referential from
which to calculate coordinates is quite unsolvable for texts that
do not mention it explicitly. The IE task could involve deeper
semantic analysis that provides means of constructing a more
global spatial representation of the scene.

    \section{Conclusion}

This paper has presented a prototype system that is able to
process correctly 17\% of our corpus of car accident reports up to
a 3D simulation of the scene. The chosen approach divides the task
between information extraction to fill templates and planning to
animate the scene. It leads to encouraging results, considering
that the information retrieval could be improved by integrating
more elaborate methods to deal with space and time in written
texts.

    \bibliographystyle{acl}
    \bibliography{references}

\end{document}